\documentclass[journal]{IEEEtran}

\IEEEoverridecommandlockouts                              

\usepackage{amsmath} 
\usepackage{amsfonts} 
\usepackage{amssymb}
\usepackage{graphicx}
\usepackage{caption}
\usepackage{multirow}
\usepackage{float}
\usepackage{color}
\usepackage{booktabs}
\usepackage{hyperref}
\usepackage[ruled,linesnumbered,lined,boxed,commentsnumbered]{algorithm2e}
\usepackage{algorithmicx}
\usepackage[numbers,sort&compress]{natbib}

\graphicspath{ {./figure/} }

\begin{document}
\captionsetup{font={small}}
\title{\LARGE \bf
Dream to Drive with Predictive Individual World Model}

\author
{
${\text{Yinfeng Gao}}^{*}$, $\text{Qichao Zhang}$, $\text{Da-Wei Ding}$,~\IEEEmembership{Senior Member, IEEE}$, $ $\text{Dongbin Zhao}$, ~\IEEEmembership{Fellow, IEEE}$ $
\thanks{
This work is supported by the National Key Research and Development Program of China under Grants 2022YFA1004000, the National Natural Science Foundation of China (Nos. 62273035, U21A20475, 62103041, 62173325), and the Science and Technology Program of Gansu Province (21ZD4GA028).
(\it{Yinfeng Gao and Qichao Zhang contributed equally to this work.})
(\it{Corresponding Author: Da-Wei Ding; Dongbin Zhao.})
} 
\thanks{Yinfeng Gao and Da-Wei Ding are with the School of Automation and Electrical Engineering, University of Science and Technology Beijing, Beijing 100083, China, and with Key Laboratory of Knowledge Automation for Industrial Processes, Ministry of Education, Beijing 100083, China. Yinfeng Gao is also with The State Key Laboratory of Multimodal Artificial Intelligence Systems, Institute of Automation, Chinese Academy of Sciences, Beijing, 100190, China (e-mail: gaoyinfeng07@gmail.com; dingdawei@ustb.edu.cn).}
\thanks{Qichao Zhang and Dongbin Zhao are with The State Key Laboratory of Multimodal Artificial Intelligence Systems, Institute of Automation, Chinese Academy of Sciences, Beijing 100190, China, and also with the School of Artificial Intelligence, University of Chinese Academy of Sciences, Beijing 100049, China (e-mail: zhangqichao2014@ia.ac.cn; dongbin.zhao@ia.ac.cn).}
\thanks{$^{*}$ This work was done when Yinfeng Gao conducted internship at CASIA.}
\thanks{Codes and pre-trained models: \href{}{https://github.com/gaoyinfeng/PIWM}.}
}

\maketitle
\markboth{IEEE TRANSACTIONS ON INTELLIGENT VEHICLES, VOL. , NO. , 2024
}{Roberg \MakeLowercase{\textit{et al.}}: High-Efficiency Diode and Transistor Rectifiers}

\begin{abstract}
It is still a challenging topic to make reactive driving behaviors in complex urban environments as road users' intentions are unknown. 
Model-based reinforcement learning (MBRL) offers great potential to learn a reactive policy by constructing a world model that can provide informative states and imagination training.
However, a critical limitation in relevant research lies in the scene-level reconstruction representation learning, which may overlook key interactive vehicles and hardly model the interactive features among vehicles and their long-term intentions.
Therefore, this paper presents a novel MBRL method with a predictive individual world model (PIWM) for autonomous driving. PIWM describes the driving environment from an individual-level perspective and captures vehicles' interactive relations and their intentions via trajectory prediction task.
Meanwhile, a behavior policy is learned jointly with PIWM. It is trained in PIWM's imagination and effectively navigates in the urban driving scenes leveraging intention-aware latent states.
The proposed method is trained and evaluated on simulation environments built upon real-world challenging interactive scenarios. Compared with popular model-free and state-of-the-art model-based reinforcement learning methods, experimental results show that the proposed method achieves the best performance in terms of safety and efficiency.
\end{abstract}


\section{Introduction}

In recent years, self-driving has drawn widespread interest from the automotive engineering and artificial intelligence communities\cite{liu2022prediction,li2024osm,ding2023Dos}. 
However, how to make human-like and reactive driving decisions\cite{wei2023humanlike} in complex urban scenarios, such as unprotected left turns at intersections\cite{chen2023humanlike}, is still an open problem, since the intentions of other road users are unknown to the autonomous vehicle (AV), and AV needs to make relatively long-term decisions in the proper timing to navigate in these scenarios safely and efficiently.
Plenty of preliminary decision-making frameworks are based on hand-designed rules\cite{ferguson2008reasoning},\cite{paden2016survey}, which have been successfully implemented\cite{leonard2008perception}, and offer high explainability thanks to their rigorous logic rules. Nonetheless, this type of approach becomes laborious as the traffic density increases, and it is challenging to cover all safety-critical cases only with manually designed rules, which results in poor generalization ability.
Fortunately, learning-based decision-making methods\cite{bronstein2022hierarchical},\cite{gulino2023waymax} have grown rapidly in recent years. Imitation learning (IL) and reinforcement learning (RL) are two types of mainstream learning-based approaches as they can directly learn decision-making policies based on expert demonstrations or interactions with the driving environment. 
On the one hand, IL methods\cite{ly2020learning} learn decision policies by mimicking expert demonstrations in a supervised learning manner\cite{hawke2020urban},\cite{bansal2018chauffeurnet}, which is efficient and scalable, but due to the mismatch of training and inference distribution, the compounding error inevitably grows and causes the distribution shift issue. 
On the other hand, RL methods\cite{li2023reinforcement} learn decision policies through online interaction with the environment, thus preventing the mismatching problem. 
Many works strive to tackle the decision-making problem utilizing model-free RL methods in typical driving scenarios, such as intersection navigation\cite{liu2022multi},\cite{yuan2021deep}, highway lane changing\cite{wang2021highway}, vehicle-pedestrian interaction\cite{crosato2023interaction}, and exploration in unknown environments\cite{li2019deep}. 

The model-free RL methods suffer from several drawbacks that reduce their applicability in challenging urban driving environments, one is that it lacks the ability to learn long-term behaviors, and the other is low data efficiency.
Model-based RL (MBRL)\cite{ha2018recurrent},\cite{kaiser2019model} provides a promising path to mitigate the aforementioned issues by introducing a learned world model, which usually builds a transition dynamic, a reward function, and optionally a representation module. 
The world model can be used to provide extra data and generate informative representation to facilitate policy training.
Dreamer is a leading method in the MBRL family, equipped with a world model called the recurrent state-space model (RSSM)\cite{hafner2019learning}. Its V1 version\cite{hafner2019dream} proposes to train a policy fully in the world model's imagination and exceeds previous model-free and model-based methods in the DeepMind Control Suite. 
Its V2 version\cite{hafner2020mastering} applies multiple modifications on V1 such as utilizing discrete representations, and realizes human-level performance on the Atari benchmark.
Recently, its V3 version\cite{hafner2023dreamerv3} has made a lot of effort on applicability and achieves state-of-the-art (SOTA) outcomes across a wide spectrum of tasks using fixed hyperparameters, showing superior performance and generality.

\begin{figure*}[htbp]
\centering
\begin{center}
\scriptsize
\includegraphics[width=16cm]{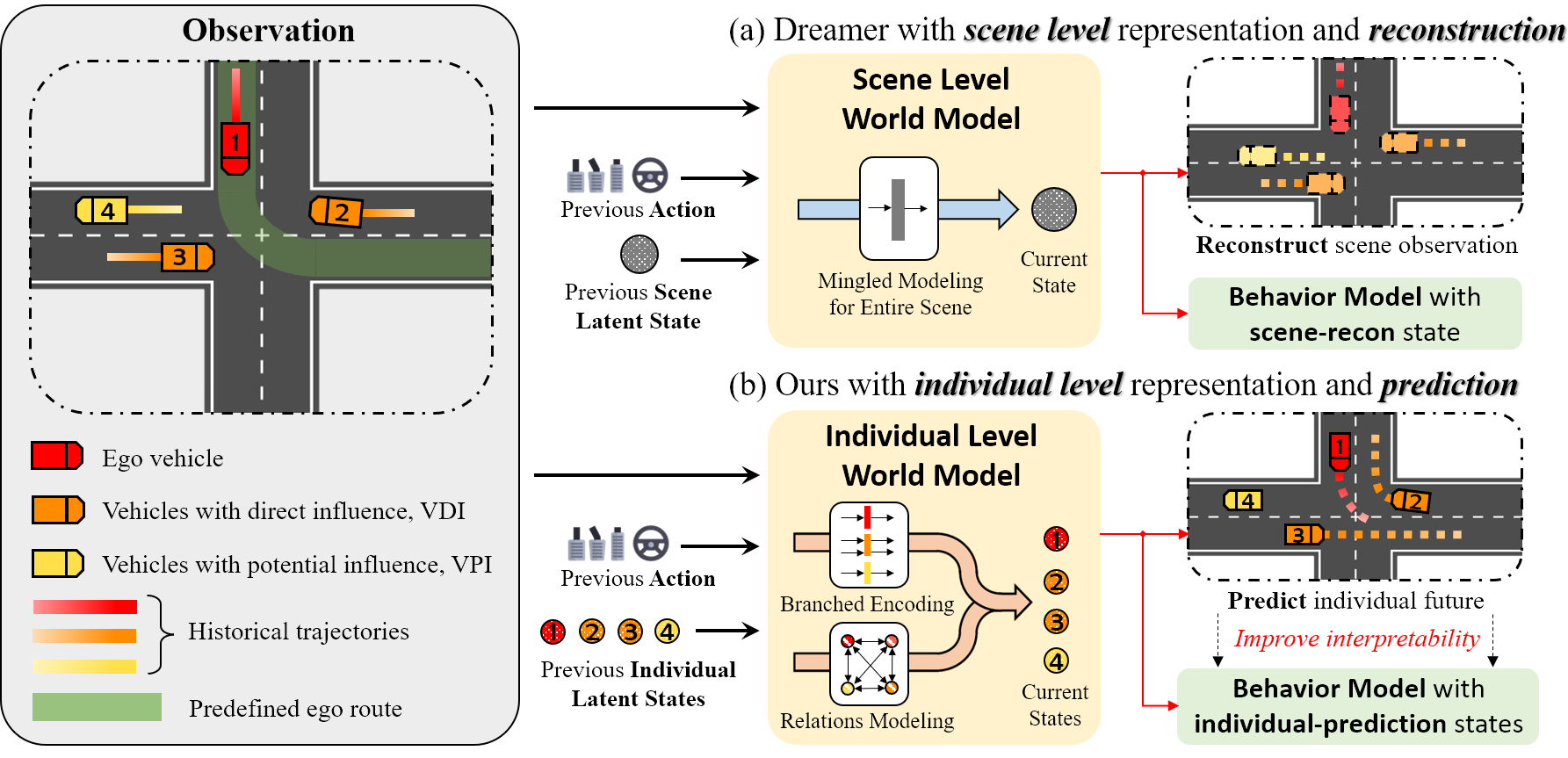}
\caption{
Comparison of Dreamer\cite{hafner2023dreamerv3} and our method.
Considering a complex driving scenario with an ego vehicle and several social vehicles. 
Dreamer learns a scene-level world model, its representational learning focuses on reconstructing the current observation. 
The behavior model learns to operate on the mingled state, depicted as a grey circle.
In contrast, our method develops the world model in an individual-level framework, where each vehicle in the scene is classified and modeled separately with branched networks and owns a unique state, symbolized by a numbered colored circle. 
We further improve the individual-level world model by explicitly modeling the relations between vehicles, and replacing the reconstruction with trajectory prediction to capture the latent intentions or motion trends of vehicles. 
}
\label{fig: Framework}
\end{center}
\vspace{-1.3em}
\end{figure*}

Some previous works have studied deploying Dreamer to the autonomous driving tasks\cite{zhang2021steadily,gao2024enhance,brunnbauer2022latent}, where the agent takes raw sensor input or bird's eye view (BEV) and outputs low-level controls like throttle and steering. 
However, there are still some challenges that make it inappropriate to directly employ Dreamer in complex driving environments.
The first challenge is that it can only describe the environment in a scene-level representation under the current framework, which means the information of different vehicles is mingled together and can not be modeled independently, resulting in high sample complexity and vague dependencies between vehicles. 
The second challenge is that the original Dreamer uses observation reconstruction for representation learning, resulting in inadequate modeling of vehicles' intentions, which is essential for making feasible decisions in interactive scenarios. Despite that Dreamer naturally learns a predictive state through training a transition dynamic, it is not enough to show vehicles' long-term intentions. 
To overcome the above challenges, we first assume that vehicle detection and localization are realized by existing perception algorithms, and use vehicles' poses in vector form as inputs. An individual world model framework is then proposed based on vector inputs, which separately models changeable numbers of vehicles at an individual level leveraging a branched network, where different branches are used to model different kinds of vehicles, e.g., ego and social vehicles. 
Note that ISO-Dream\cite{pan2022iso} also conducts a branched world model, but their purpose is to decouple controllable and uncontrollable states from fixed-size visual inputs, which is different from ours.
Furthermore, to capture vehicles' interactive relations and learn their long-term intentions, we introduce the interactive prediction task to the individual world model, replacing the original observation reconstruction task. 
Specifically, we enhance the transition dynamic of the individual world model with a self-attention mechanism that operates on the individual features of the vehicles. The vehicle-to-vehicle interactive relations are finely modeled with self-attention, and the vehicles' relations are represented by the attention values.
The states of vehicles are formed by concatenating attention values with their individual features. These states are further used to predict vehicles' future trajectories. By learning to predict vehicles' future trajectories, vehicles' long-term intentions can be encoded in their latent states,
finally improving the performance of the world model and the behavior policy.

Overall, We present a novel model-based RL driving method for complex urban driving with a designed \textbf{p}redictive \textbf{i}ndividual \textbf{w}orld \textbf{m}odel (\textbf{PIWM}).
The main contributions of this paper are summarized as follows:
\begin{itemize}
\item We present a novel model-based RL framework for autonomous driving. The world model learns to encode driving scenes from an individual perspective by leveraging branched networks, which helps to reduce sample complexity and improve decision performance in complex scenarios.
\item We enhance the transition dynamic of the individual world model by explicitly modeling the interactive relations between vehicles. The trajectory prediction task is introduced to learn informative representations, which derive vehicles' intentions and benefit the downstream decision-making task.
\item The proposed method is validated on hundreds of driving scenarios imported from the real-world driving dataset, experimental result shows that our method achieves conspicuous advantages in sample efficiency and converged performance compared with existing RL methods,
gaining $18.81\%$ boost on success rate in the large-scale experiment compared with SOTA MBRL method DreamerV3.
Examples of decoded predictive trajectories are illustrated to improve the interpretability of the behavior further.
\end{itemize}


\section{Related Work}
\subsection{Reinforcement Learning in Autonomous Driving}
Reinforcement learning (RL) is a popular approach for tackling decision-making problems in various domains. It tries to learn an optimal policy that maximizes the expected accumulated rewards by interacting with the environment and learning from trial and error. Plenty of previous works study the autonomous driving problem leveraging RL from different aspects, \cite{sallab2017deep} proposes an RL-based framework which incorporates Deep Q learning (DQN)\cite{mnih2015human} with attention mechanism and performs well in lane-keeping tasks, \cite{nishitani2020deep} proposes a vehicles controller based on DQN and effectively navigate in merging scenarios, \cite{wurman2022outracing} extends soft actor-critic (SAC)\cite{haarnoja2018soft} and defeats best human drivers in racing games.
RL is also used in other driving-related tasks, such as traffic signals control\cite{mao2023signal}, cooperative driving\cite{zhang2023bilevel}, traffic generation\cite{zhang2022trajgen}, and so on. 
A prominent problem for RL is the low sample efficiency, which stunts its implementation to the real driving task. To improve sample efficiency, a practicable way is to make use of an imitative export policy to regularize the RL agent's behavior\cite{huang2022efficient}, or introduce auxiliary tasks to learn informative state representations for autonomous driving tasks\cite{chen2021interpretable},\cite{kargar2022increasing}. 

\subsection{Model-based RL in Autonomous Driving}
Model-based methods learn skills by establishing a world model, which has been primarily explored in the context of RL\cite{luo2024survey}.  
Some works have studied learning driving behaviors in an MBRL manner, \cite{wu2022uncertainty} uses an ensemble of world models to evaluate the uncertainty of the generated samples, providing reliable augmented experiences to accelerate the training process. Wang. et al. introduce a method that offers to adaptively select the roll-out horizon of the differentiable world model to provide a more accurate value estimation, it learns lane-changing maneuver from BEV semantics\cite{wang2022dynamic}. 
Learn to drive with virtual memory (LVM)\cite{zhang2021steadily} and semantic masked recurrent world model (SEM2)\cite{gao2024enhance} are both built on Dreamer, where LVM designs a double critic structure for the behavior model and shows a more stable learning process in the lane-keeping task. SEM2 enhances the original Dreamer structure by learning BEV semantics from raw sensor inputs, it is deployed to control the AV effectively in urban scenarios. ISO-Dream\cite{pan2022iso} learns a world model that separates the visual dynamics into controllable and uncontrollable states, and subsequently, a vehicle control policy is trained on the disentangled states.
On the other hand, Model-based imitation learning (MBIL)\cite{hu2022model},\cite{wang2023drivedreamer} combines imitation learning and world model learning, which can directly learn a driving policy from an offline dataset without online interaction with the environment.

Learning a world model is fundamental for the AV to comprehend the environment and make decent decisions in complex driving scenarios. However, the model-based approaches discussed above mainly consider pixel-level inputs, learning a scene-level world model that can not clearly model vehicles' motions and is not suitable for learning sophisticated interactive maneuvers. 
This paper uses effective vector-level inputs borrowed from SOTA motion forecasting methods\cite{sun2022m2i,gu2021densetnt,li2023pih}. 
Furthermore, we propose a novel MBRL framework for autonomous driving, in which the world model is designed to model the driving scene from an individual perspective. 
Consequently, the vehicles' motions can be finely modeled separately and the sample complexity is reduced, finally improving driving performance.

\subsection{Join Prediction and Planning for Interactive Modeling}
Crosato. et al. systematically and comprehensively review the current state of interactive-aware behavior modeling and motion planning in the context of AV\cite{crosato2023social}.
To consider future interactive modeling, the joint prediction and planning method has been investigated for learning-based reactive driving policy in complex urban environments \cite{hagedorn2023rethinking}.
Huang. et al. present a differentiable integrated prediction and planning (DIPP) framework, which jointly optimizes the prediction and planning tasks from the prediction and optimization perspective\cite{huang2023differentiable}. 
\cite{kargar2022increasing} tries to integrate motion prediction with RL and trains a trajectory prediction module in BEV form, which serves as a safety indicator to prevent possible collisions. 
From the game model perspective, \cite{sarkar2021solution} investigates naturalistic driving behavior as a bounded rational activity through a behavioral game-theoretic lens. 
\cite{sarkar2022generalized} develops a framework of generalized dynamic cognitive hierarchy for both modeling naturalistic human driving behavior as well as behavior planning for AV. 
Inspired by the level-k game theory, Gameformer\cite{huang2023gameformer} provides hierarchical game-theoretic modeling and learning of Transformer-based interactive prediction and planning. 
Note that those popular game models are very similar to the imagination process of MBRL, which both focus on modeling the future interactive reasoning for planning. The main difference is that game models are about interactive modeling explicitly with stronger interpretability, while the imagination of MBRL is about interactive modeling in the latent space with higher efficiency.

Recently, large language models (LLMs) have shown promise in autonomous driving, due to their strong reasoning capabilities and generalization potential. For example, GPT-Driver\cite{mao2023gpt} represents the planner inputs and outputs as language tokens, and leverages the LLMs to predict driving trajectories through a language description of coordinate positions. Dilu\cite{wen2023dilu} combines a Reasoning and a Reflection module to enable the LLMs to perform decision-making based on common-sense knowledge and evolve continuously. LMDrive\cite{shao2023lmdrive} provides a novel language-guided, end-to-end, closed-loop autonomous driving framework. In those works, the future interactive modeling is taken into account by the LLMs, featuring a designed chain-of-though output. However, the process of reasoning is very time-consuming for LLMs, how to combine the reasoning ability of LLMs with the efficient imagination ability of world models will be a very attractive research topic for autonomous driving.


\begin{figure*}[htbp]
\centering
\includegraphics[width=18cm]{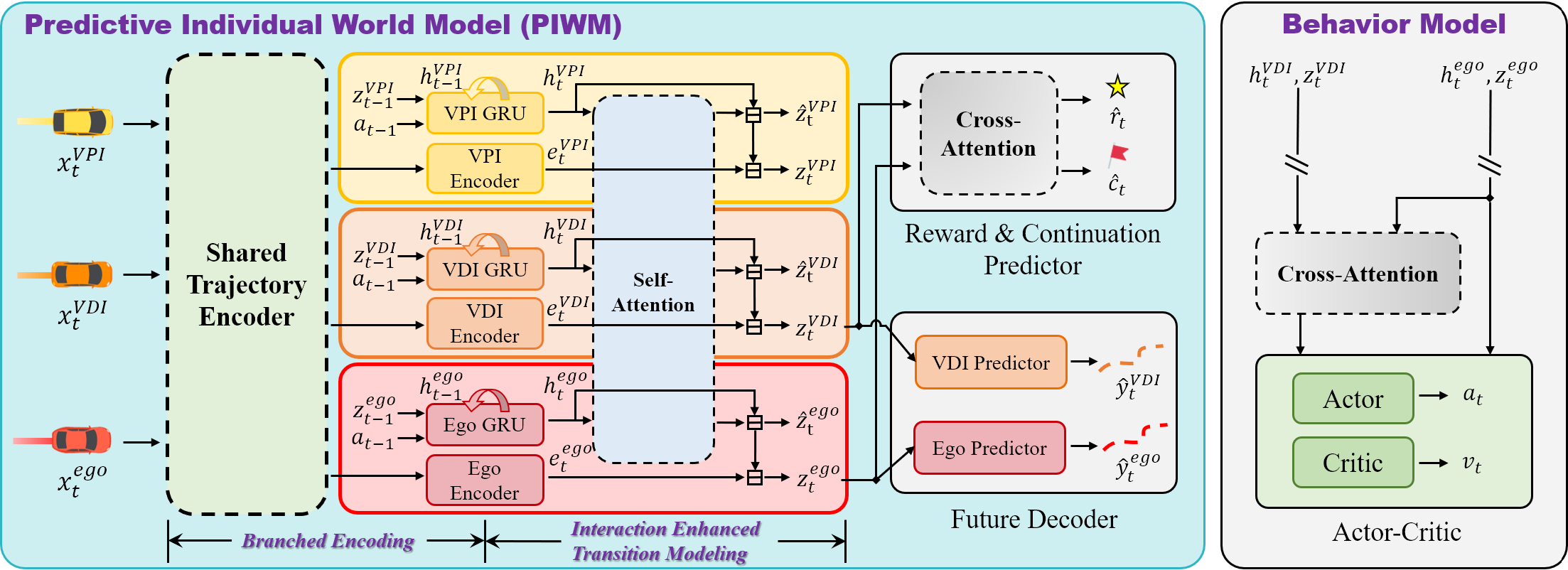}
\caption{The detailed structure of PIWM and Behavior Model. The modules represented by solid lines are branched-only, the other modules represented by dash lines are shared between branches. Note that the gradients of actor and critic are stopped from flowing backward through latent states, which makes the representation learning purely happen in the world model learning phase. Network modules are all formed as multi-layer perceptrons (MLPs) since no image observations are considered.}
\label{fig: detailed sturcture}
\vspace{-1.3em}
\end{figure*}

\section{Method}
In this section, we introduce the technical details of our method, which jointly learns a world model that describes the driving environment in latent space and a behavior model that makes driving decisions. 
Similar to DreamerV3\cite{hafner2023dreamerv3}, the world model is learned from experience data, and the behavior model is trained in the world model's imagination, which is then employed in the real environment to collect more experience data.

An overview of our method is illustrated in Fig. \ref{fig: Framework}(b). 
For a complex driving scenario, the ego vehicle's observation contains historical trajectories of itself and surrounding social vehicles. 
Based on the observation, vehicles' previous latent states, and ego's previous action, our proposed predictive individual world model (PIWM) can extract both vehicles' individual features and their interactive relations, which are concatenated together to form vehicles' current latent states. The latent states are learned to forecast vehicles' future trajectories and capture vehicles' intentions. 
The behavior model can make reactive decisions based on these intention-aware states.
In the following, we first define the decision-making problem in section \ref{section: problem formulation}. Then we elaborate on the structure of PIWM and the behavior model in section \ref{section: PIWM} and section \ref{section: behavior} respectively. Finally, the learning process of the method and the pseudo-code are given in \ref{section: learning process}.

\subsection{Problem Formulation}
\label{section: problem formulation}
Since the intentions of social vehicles cannot be directly observed by the AV, we consider the decision-making problem as a partially observable Markov decision process (POMDP), which can be defined using a tuple $(\mathcal{X}, \mathcal{S}, \mathcal{A}, P, O, R, \gamma)$, where $\mathcal{X}$, $\mathcal{S}$, and $\mathcal{A}$ are the observation space, state space, and action space, respectively. $P(s_{t-1}, a_{t-1}, s_t)=p(s_t|s_{t-1}, a_{t-1})$ is the transition dynamic which maps the state $s_{t-1}$ and action $a_{t-1}$ to a probability distribution of the next states $s_{t}$, and $O(s_t, x_t)=p(s_t|x_t)$ is the observation function which maps the observation $x_t$ to state $s_t$. $\mathcal{R}: \mathcal{S} \times \mathcal{A} \to \mathbb{R} $ is the reward function that maps the state and its corresponding action to a scalar reward value, and $\gamma \in (0, 1)$ is the discount factor. POMDP aims to find an optimal policy $\pi^*: \mathcal{S} \to \mathcal{A}$, maximizing the expected discounted cumulative rewards, i.e. return. 

Considering a driving scenario that consists of an ego vehicle and $N$ surrounding social vehicles, the observation at time $t$ can be written as $x_t = \{x_t^{ego}, x_t^{1},...,x_t^{N}\} \in \mathcal{X}$, where $x_t^{ego}$ and $x_t^{i}$ represent the ego and $i$th vehicle's historical $2s$ positions and headings that can be observed from the perception module. Similarly, the state at time $t$ can be defined as $s_t=\{s_t^{ego}, s_t^{1},...,s_t^{N}\} \in \mathcal{S}$, where $s_t^{ego}$ and $s_t^{i}$ reveal the ego and $i$th vehicle's latent information including driving intentions that cannot be observed directly. We consider the action at time $t$ as $a_t \in \mathcal{A}$, which is a route-conditional longitudinal speed control, i.e., the route is the desired path the ego should follow, which is usually derived from the navigation system, and $a_t$ controls the ego’s target speed on the route. Note that action space $\mathcal{A}$ is discrete in this work.
More details about the designs of the observation space and action space are further discussed in section \ref{section: ES}.

\subsection{Predictive Individual World Model} 
\label{section: PIWM}
The Predictive Individual World Model (PIWM) is particularly designed for autonomous driving.
It learns to model the driving environment in latent space from an individual perspective and obtains informative representations for the behavior model through interactive prediction. 
Note that "predictive" in the name not only reflects PIWM's ability to learn the transition dynamic $s_t \sim p(s_t|s_{t-1}, a_{t-1})$, but also indicates that vehicles' latent states $s_t$ are learned from the interactive prediction task. The architecture of PIWM is shown in the left part of Fig. \ref{fig: detailed sturcture} and the details are described below.

\textbf{Branched Encoding.}
In the PIWM, we divide social vehicles into two categories based on their distances to the ego vehicle, and simply assume that vehicles that are relatively close to the ego are more likely to have a direct influence on its decision, similar to DIPP\cite{huang2023differentiable}. The other vehicles that are relatively far away from the ego may have an indirect influence on the ego's decision through their interactions with the closer ones, or no influence at all. We name them as vehicles with direct influence (VDI) and vehicles with potential influence (VPI), respectively. Accordingly, the observation at time $t$ can be redefined as $x_t=\{x_t^{ego}, x_t^{VDI}, x_t^{VPI}\}$, where $x_t^{VDI}=\{x_t^{d_i}\}_{i \in [ 1, N_d ]}$ and $x_t^{VPI}=\{x_t^{p_j}\}_{j \in [ 1, N_p ]}$. The state at time $t$ can be redefined as $s_t=\{s_t^{ego}, s_t^{VDI}, s_t^{VPI}\}$. Note that $N_{d}$ and $N_{p}$ are numbers of the VDI and VPI and $N=N_{d}+N_{p}$.
As shown Fig. \ref{fig: detailed sturcture}, we extract vehicles' embedding $e_t=\{e_t^{ego},e_t^{VDI},e_t^{VPI}\}$ from their observation $x_t=\{x_t^{ego},x_t^{VDI},x_t^{VPI}\}$ leveraging branched networks. Specifically, the observations of vehicles including the ego, VDI, and VPI are processed using a shared trajectory encoder to extract common knowledge first, and then passed through subsequently branched encoders:
\begin{equation}
\label{eq: et}
\begin{split}
\begin{array}{ll}
e_t^{ego}=\texttt{Enc}_{\theta_{1}}(\texttt{Enc}_{\phi}(x_t^{ego})) \\ [4pt]
e_t^{VDI}=\texttt{Enc}_{\theta_{2}}(\texttt{Enc}_{\phi}(x_t^{VDI})) \\ [4pt]
e_t^{VPI}=\texttt{Enc}_{\theta_{3}}(\texttt{Enc}_{\phi}(x_t^{VPI})) \\ 
\end{array}
\end{split}
\end{equation}
where $\phi$ is the parameters of the shared trajectory encoder and $\theta_{1}, \theta_{2}, \theta_{3}$ are the parameters of the ego encoder, VDI encoder, and VPI encoder respectively.

\textbf{Interaction Enhanced Transition Modeling. }
The design of the transition model follows the RSSM architecture\cite{hafner2019learning}, where the latent states $s_t$ are composed of the deterministic states $h_t$ and the stochastic states $z_t$, i.e. $s_t=\{h_t, z_t\}$, and gated recurrent unit (GRU) cells are used to keep the historical information. 
In this work, we additionally consider modeling the vehicles' interactive relations within the latent states, which enhances the transition model by explicitly capturing vehicles' inner connections and providing important features for the following trajectory prediction task.
We first derive the deterministic states $h_t$ and obtain such interactive relations $self\_att_t$ by using the self-attention mechanism:
\begin{equation}
\label{eq: ht}
\begin{split}
\begin{array}{ll}
h_t^{ego}=\texttt{GRU}_{ego}(h_{t-1}^{ego}, z_{t-1}^{ego}, a_{t-1}) \\ [4pt]
h_t^{VDI}=\texttt{GRU}_{VDI}(h_{t-1}^{VDI}, z_{t-1}^{VDI}, a_{t-1}) \\ [4pt]
h_t^{VPI}=\texttt{GRU}_{VPI}(h_{t-1}^{VPI}, z_{t-1}^{VPI}, a_{t-1}) \\ [4pt]
self\_att_t=\texttt{Self\_Attention}(h_t^{ego}, h_t^{VDI}, h_t^{VPI}) \\ 
\end{array}
\end{split}
\end{equation}
where the GRU cells are also designed in a branched style, and $\texttt{Self\_Attention}$ represents the self-attention mechanism in which the queries, keys, and values (\textbf{Q, K, V}) are the encoded vehicles' deterministic states by processing them with linear projections $L_q^{self}$, $L_k^{self}$, and $L_v^{self}$.

Then the process of getting prior and posterior of the stochastic states can be written as:
\begin{equation}
\label{eq: prior&post}
\begin{split}
\begin{array}{ll}
\bar{z}_t \sim p(\bar{z}_t \mid h_t, self\_att_t) \\ [4pt]
z_t \sim q(z_t \mid h_t, self\_att_t, e_t) \\
\end{array}
\end{split}
\end{equation} 
here we use $\bar{z}_t$ to denote the prior and $z_t$ to denote the posterior, their difference lies in whether have access to the embeddings. The stochastic states are also encoded and sampled using branched networks as before, we omit the branches for simplification. 
Note that distances between ego and other vehicles change when interacting with the environment, the members of VDI and VPI are constantly changing accordingly. 
For VDI or VPI which are newly added to their category, their latent states are initialized with zeros.

\textbf{Future Decoding. } 
The decoders of PIWM are designed to predict vehicles' future trajectories from the latent states $s_t$, making $s_t$ more informative of vehicles' intentions or their motion trends. It should be clarified that the decoder has only two branches to predict future trajectories of ego and VDI:
\begin{equation}
\begin{split}
\begin{array}{ll}
\hat{y}_t^{ego}=\texttt{Dec}_{\varphi_1}(s_t^{ego}) \\ [4pt]
\hat{y}_t^{VDI}=\texttt{Dec}_{\varphi_2}(s_t^{VDI}) \\
\end{array}
\end{split}
\end{equation} 
where $\varphi_1$ and $\varphi_2$ are the parameters of the ego decoder and VDI decoder, and $\hat{y}_t$ represents the predicted trajectories. 

We do not set another branch decoder to predict VPI's future for two reasons. The first reason is that the whole framework is designed as decision-oriented, therefore we tend to learn state representations that can better describe ego and VDI which have a direct influence on the ego's decisions, VPI's states are mainly used to portray their potential influence to the ego and VDI through interactive relations modeling, bringing in VPI's future decoder may mislead the learning purpose of VPI's states. The second reason is that our method is fundamentally an online RL method and VPI does not always exist in varying driving scenarios, which results in relatively insufficient data collection to train a proper future decoder for VPI.

\textbf{Reward and Continuation Predicting. } 
The reward and continuation prediction modules are used to predict reward value $r_t$ and continuation flag $c_t$ given latent states $s_t$. 
Particularly, we don't consider VPI's states directly in input (which are indirectly considered in the latent states of the ego and VDI through interactive relations modeling) and utilize a cross-attention (ego-attention) \cite{leurent2019social} mechanism to fuse the states of ego and VDI:
\begin{equation}
\label{eq: reward&continue}
\begin{split}
\begin{array}{ll}
cross\_att_t=\texttt{Cross\_Attention}(s_t^{ego}, s_t^{VDI}) \\ [4pt]
\hat{r}_t \sim p(\hat{r}_t \mid s_t^{ego}, cross\_att_t) \\ [4pt]
\hat{c}_t \sim p(\hat{c}_t \mid s_t^{ego}, cross\_att_t) \\
\end{array}
\end{split}
\end{equation}
where $\hat{r}_t$ and $\hat{c}_t$ are the predicted reward value and continuation flag, $\texttt{Cross\_Attention}$ is the cross-attention mechanism, here query ($\textbf{Q}$) is the encoded ego's state with linear projection $L_q^{cross}$, keys ($\textbf{K}$) and values ($\textbf{V}$) are the encoded ego's and VDI's states with linear projections $L_k^{cross}$ and $L_v^{cross}$. Different from the self-attention used in the transition model, cross-attention uses only the ego's state to generate the query, which better fits ``ego-centric'' tasks like reward value prediction and continuation flag prediction.

\subsection{Behavior Generation} 
\label{section: behavior}
An actor-critic method is adopted as the behavior model following DreamerV3\cite{hafner2023dreamerv3}. The structure of the behavior model is shown in the right part of Fig. \ref{fig: detailed sturcture}, which operates on the ego's state $s_t^{ego}$ and VDI's states $s_t^{VDI}$, output action with the actor and value estimation with the critic. Since we also view behavior generation as an ``ego-centric'' task, the cross-attention $cross\_att_t=\texttt{Cross\_Attention}(s_t^{ego}, s_t^{VDI})$ is used again for state fusing. The actor and critic models are defined as:
\begin{equation}
\label{eq: actor&critic}
\begin{split}
\begin{array}{ll}
\text{Actor:}& a_t \sim \pi_\omega(a_t \mid s_t^{ego}, cross\_att_t) \\ [4pt]
\text{Critic:}& v_t \sim u_\psi(s_t^{ego}, cross\_att_t) \approx \mathbb{E}_{\pi_\omega}[R_t] \\
\end{array}
\end{split}
\end{equation}
where $u$ represents the critic, $\omega$ is the parameters of actor and $\psi$ is the parameters of critic, $R_t$ is the return.
Note that for efficient convergence considering potentially widespread return distribution, the critic is designed to learn a discrete distribution\cite{bellemare2017distributional}, i.e. twohot softmax distribution, as in DreamerV3\cite{hafner2023dreamerv3}.
According to section \ref{section: problem formulation}, the action space is discrete. Hence the output of the actor is designed as a onehot distribution.

\subsection{Learning Process}
\label{section: learning process}
\textbf{World Model Learning. }
The main purpose of PIWM is to learn a meaningful state representation and model the transition between states from past experience. 
In this work, we achieve the former part through the vehicles' interactive prediction task.
Thus, we additionally derive the ground truth of the prediction trajectories $y_t$ from the data. Note that $y_t=\{y_t^{ego}, y_t^{VDI}\}$ as we don't predict VPI's future. The latter part is achieved by minimizing the KL divergence between $p(\bar{z}_t | h_t, self\_att_t)$ and $q(z_t | h_t, self\_att_t, e_t)$ in Eq. (\ref{eq: prior&post}). Given a sequence of training data $(x_t, y_t, a_t, r_t, c_t)_{t=1}^T$ sampled from the replay buffer, PIWM is optimized using the following loss functions:

\begin{equation}
\label{eq: PIWM loss}
\begin{split}
\mathcal{L}_\text{PIWM} = &\mathbb{E} \ \{ \sum_{t=1}^{T}
 \underbrace{-\ln p(y_{t}^{ego} \mid s_{t}^{ego}) -\ln p(y_{t}^{VDI} \mid s_{t}^{VDI})}_{\text {prediction log loss}} \\ 
 & \underbrace{-\ln p(r_{t} \mid s_{t}^{ego}, s_{t}^{VDI})}_{\text {reward log loss}} \underbrace{-\ln p(c_{t} \mid s_{t}^{ego}, s_{t}^{VDI})}_{\text {continuation log loss}} \\
 & \underbrace{+ \beta \mathrm{KL}[q(z_{t}^{ego} \mid h_{t}, x_{t}^{ego}) \Vert p(\bar{z}_{t}^{ego} \mid h_{t})]}_{\mathrm{KL} \text { divergence in ego's branch}} \\
 & \underbrace{+ \beta \mathrm{KL}[q(z_{t}^{VDI} \mid h_{t}, x_{t}^{VDI}) \Vert p(\bar{z}_{t}^{VDI} \mid h_{t})]}_{\mathrm{KL} \text { divergence in VDI's branch}} \\
 & \underbrace{+ \beta \mathrm{KL}[q(z_{t}^{VPI} \mid h_{t}, x_{t}^{VPI}) \Vert p(\bar{z}_{t}^{VPI} \mid h_{t})]}_{\mathrm{KL} \text { divergence in VPI's branch}}
 \}
\end{split}
\end{equation}
where $\beta$ is the hyper-parameter to scale KL losses, we set it to 0.5 as default in DreamerV3\cite{hafner2023dreamerv3}. 

\textbf{Behavior Model Learning. }
The behavior model updates its actor and critic purely from imagined sequences generated by PIWM.
These imagined sequences start from posterior states computed during PIWM training and are completed through the interaction between the actor and PIWM. We keep the members of VDI and VPI the same as in the first state during imagination.
Assuming an imagined sequence $(\bar{s}_t, \hat{a}_t, \hat{r}_t, \hat{c}_t)_{t=1}^H$, where $H=15$ is the imagination horizon,
the return targets are calculated as $\lambda$-returns $R_t^{\lambda}$ that consider rewards beyond $H$ and balance bias and variance\cite{sutton2018reinforcement}:

\begin{equation}
R_t^{\lambda} \approx \hat{r}_t + \gamma \hat{c}_t \left\{
\begin{aligned}
\left((1-\lambda)u_\psi(\bar{s}_{t+1}) + \lambda R_{t+1}^\lambda \right),& \quad t < H \\
u_\psi(\bar{s}_H),& \quad t = H
\end{aligned}
\right.
\end{equation}
where $\lambda=0.95$ is the hyper-parameter that balances long-term targets and short-term targets.

On the one hand, the actor aims to learn a policy that leads to states with the highest $\lambda$-return, whose loss function can be written as:
\begin{equation}
\label{eq: actor loss}
\begin{split}
\mathcal{L}_\text{actor} = \sum_{t=1}^{H-1} [ \operatorname{sg}(R_t^{\lambda}) / \max(1, \textit{S}) ] \\
- \eta\mathrm{H} [ \pi_\omega(a_t \mid s_t^{ego}, cross\_att_t) ]
\end{split}
\end{equation}
where $\operatorname{sg}(\cdot)$ stops the gradients. The gradients of the first term are estimated by Reinforce\cite{williams1992simple} for discrete actions. Note that $\textit{S}$ is the return scale, scaling down large returns can effectively stabilize the scale of returns\cite{hafner2023dreamerv3}. 
The second term is the entropy term that encourages exploration, $\eta = 1 \cdot 10^{-3}$ is the entropy scale.

On the other hand, the critic aims to estimate the $\lambda$-return under the current policy accurately. 
Specifically, as discussed in section \ref{section: behavior}, the critic is trained in a discrete regression approach. Therefore the return targets $R_t^{\lambda}$ are transformed into soft labels with twohot encoding\cite{schrittwieser2020mastering}:
\begin{equation}
\operatorname{twohot}(x)_i \approx \left\{
\begin{aligned}
| b_{k+1} - x | \,/\, | b_{k+1} - b_{k} |, & \ \text{if} \ i=k \\
| b_{k} - x | \,/\, | b_{k+1} - b_{k} |, & \ \text{if} \ i=k+1 \\
0, & \ \text{else} 
\end{aligned}
\right.
\end{equation}
where $k \approx \sum_{j=1}^B \operatorname{1}_{(b_j < x)}$ and $B \approx \begin{bmatrix} -20 ... +20 \end{bmatrix}$ is a sequence of equally spaced buckets $b_i$. The loss function of the critic can be written as:
\begin{equation}
\label{eq: critic loss}
\begin{split}
\mathcal{L}_\text{critic} = -\sum_{t=1}^{H-1} \operatorname{sg}(\operatorname{twohot}(R_t^{\lambda}))^T \ln p_{\psi}(b_i \mid s_t^{ego}, cross\_att_t)
\end{split}
\end{equation}
where 
$\ln p_{\psi}(b_i \mid s_t^{ego}, cross\_att_t)$ is the output of the critic network in the form of twohot softmax probability. The same discrete regression approach is also deployed in the reward prediction module.

Finally, the pseudo-code of the proposed method is given in Alg. \ref{alg: D2D}, the process of training our method can be roughly divided into two parts, one part is collecting data through online interactions with the environment, the other part is training the world model and behavior model using the collected data.

\begin{algorithm*}
  \caption{Dream to Drive with Predictive Individual World Model}
  \label{alg: D2D}
    \textbf{Hyper-parameter:} $H$: Imagination horizon, $d_{train}$: Training frequency, $ITER$: Max interaction times \\
    \textbf{Input:} Vehicles' historical trajectories in the scenario  \\
    \textbf{Output:} Ego's longitudinal target speed \\
    Initialize the replay buffer $\mathcal{B}$ using random policy and networks with random parameters. \\
    Initialize an empty temporal buffer $\mathcal{B}_{ep}$ for online-interaction data, $x_{1} \leftarrow \texttt{env.reset()}$ \\ 
    \For{{iterations} $iter=1 \ \KwTo \ ITER$}{
        \texttt{// Models training} \\
        \If {$ {\rm not} \ iter \ {\rm mod} \ d_{train}$}{
            \texttt{// PIWM training} \\
            Sample a batch of experience ${(x_t, y_t, a_t, r_t, c_t)}_{t=1}^T$ from $\mathcal{B}$ \\
            Obtain vehicles' latent states $s_t$ using Eq. (\ref{eq: et} $\sim$ \ref{eq: prior&post}) given $x_{t}$\\
            Update the world model's parameters by optimizing the loss function described in Eq. (\ref{eq: PIWM loss}) \\
            \texttt{// Behavior model training} \\
            Initialize an empty temporal buffer $\mathcal{B}_{img}$ for imagined sequences and $\bar{s}_t \gets s_t$ \\
            \For{{\rm imagined step} $i=t \ \KwTo \ t+H$}{
                    Imagine an action ${a_i}$ using the Actor in Eq. (\ref{eq: actor&critic}) given $\bar{s}_i$ \\
                Predict the corresponding reward value $\hat{r}_i$ and continuation flag $\hat{c}_i$ using Eq. (\ref{eq: reward&continue}) \\
                Collect imagined transitions $(\bar{s}_{i}, \hat{a}_i, \hat{r}_{i}, \hat{c}_{i})$ in $\mathcal{B}_{img}$ \\
                Imagine next latent states $\bar{s}_{i+1}$ with prior $\bar{z}_{i+1}$ using Eq. (\ref{eq: ht}) and Eq. (\ref{eq: prior&post}) \\
            }
            Update the Actor and Critic with Eq. (\ref{eq: actor loss}) and Eq.(\ref{eq: critic loss}) using $\mathcal{B}_{img}$ \\
        }
        \texttt{// Environment interaction} \\
        Obtain vehicles' latent states $s_{iter}$ with posterior $z_{iter}$ using Eq. (\ref{eq: et} $\sim$ \ref{eq: prior&post}) given $x_{iter}$\\
        Output an action $a_{iter}$ using the Actor in Eq. (\ref{eq: actor&critic}) given $s_{iter}$ \\
        $x_{iter+1},r_{iter},c_{iter} \leftarrow \texttt{env.step}(a_{iter})$ \\
        Store $(x_{iter}, a_{iter}, r_{iter}, c_{iter})$ in $\mathcal{B}_{ep}$ \\
        \If {$ c_{iter} $ }{
            Derive ground truth of prediction trajectories $(y_{k})_{k=iter-L}^{iter}$ from $\mathcal{B}_{ep}$, where $L$ is the length of $\mathcal{B}_{ep}$ \\
            Store experience with ground truth prediction $(x_k, y_k, a_k, r_k, c_k)_{k=iter-L}^{iter}$ in $\mathcal{B}$ \\
            Empty $\mathcal{B}_{ep}$ for a new episode \\ 
            $x_{iter+1} \leftarrow \texttt{env.reset}()$ \\
        }
        
    }
\end{algorithm*}


\section{Experiments}

\subsection{Experiment Setup}
\label{section: ES}
\subsubsection{\textbf{Dataset}}

\begin{figure}[htbp]
    \centering
    \includegraphics[width=8.7cm]{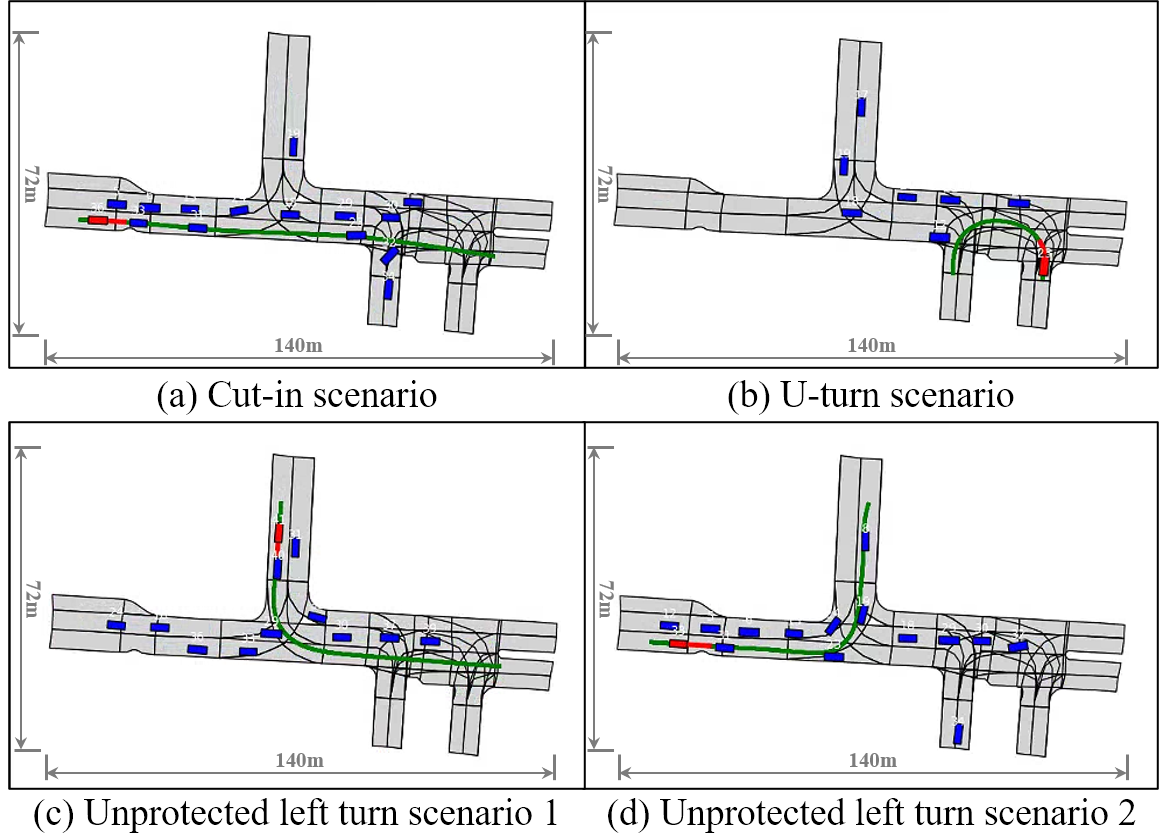}
\caption{Typical scenarios of the experiments. Where the ego vehicle is \textcolor{red}{\textbf{red}} and other social vehicles are \textcolor{blue}{\textbf{blue}}. The \textcolor[RGB]{23,104,21}{\textbf{green line}} indicates the ego's predefined route and the \textcolor{red}{\textbf{red line}} indicates the ego's heading.}
\label{fig: 4 typical scenarios}
\vspace{-1.3em}
\end{figure}

To evaluate the decision-making performance of our method, we conduct experiments on the INTERACTION dataset \cite{zhan2019interaction}, which is a large-scale real-world dataset including 16.5h highly interactive and diverse scenarios.
In this work, we focus on the challenging decision-making problem in unsignalized intersection scenarios. 
Specifically, we choose the map and track records of $\texttt{DR\_USA\_Intersection\_EP0}$ from the INTERACTION dataset. The map mainly features a horizontal, bidirectional main road that spans approximately 140 meters, which includes three road intersections without traffic signals. The total length of the roads running in the vertical direction is approximately 72 meters. 
The track records include 732 vehicles’ recorded scenarios, in which the typical scenarios are shown in Fig. \ref{fig: 4 typical scenarios}, covering the cut-in scenario where ego needs to choose the appropriate timing to merge into the traffic flow, the U-turn and unprotected left turn scenarios where ego needs to identify others' intentions to make interactive decisions.

The agent is both trained and evaluated by replacing the recorded vehicles with ego vehicles. When choosing egos, we further filter out vehicles that have large shapes ($length \textgreater 5.5m$), brief durations($\textless 5s$), or short tracks($\textless 20m$), and finally obtain 640 meaningful vehicles’ recorded scenarios. 
Note that the excluded vehicles still run as background vehicles in the scenarios.

\begin{figure}[htbp]
    \centering
    \includegraphics[width=8.5cm]{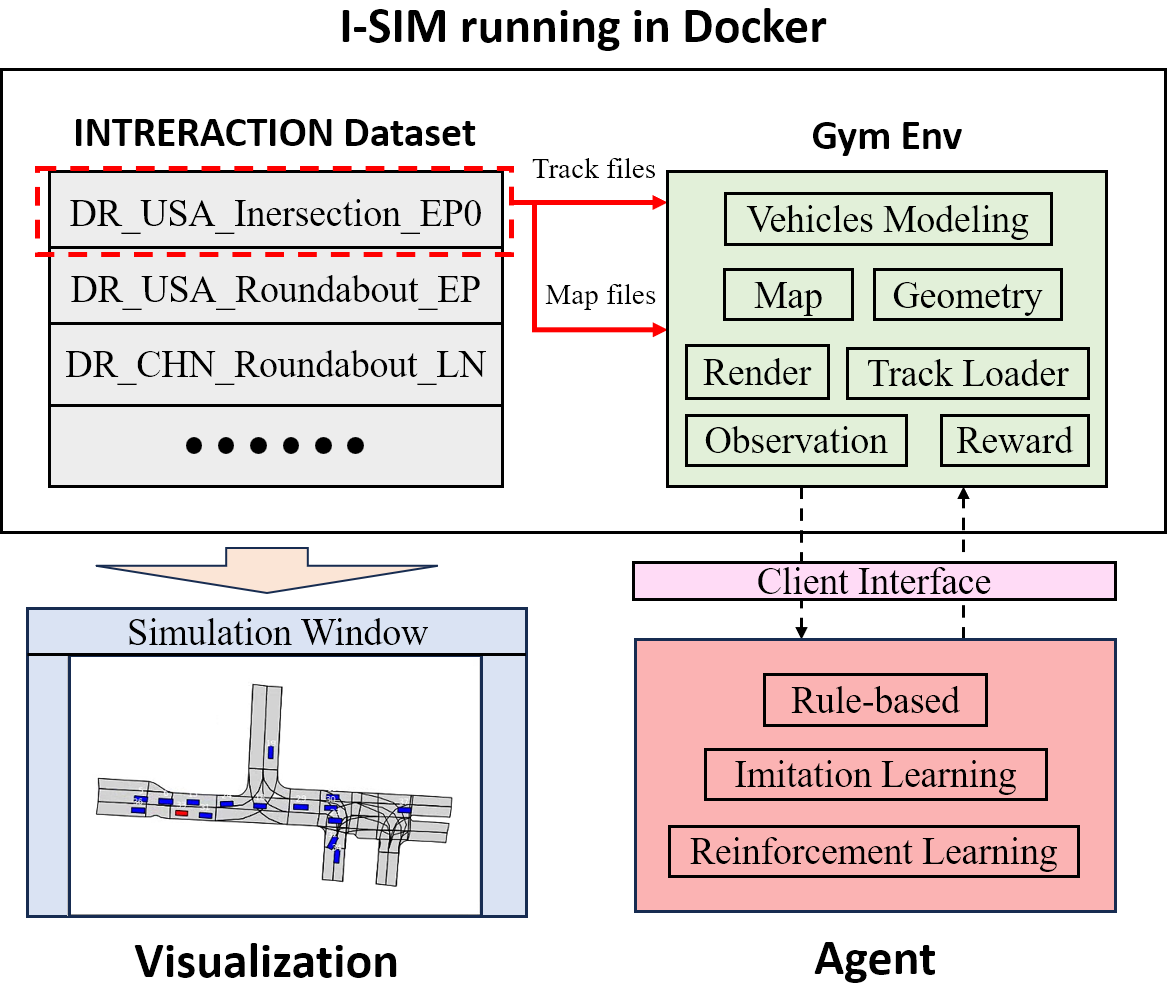}
\caption{The workflow of the I-SIM simulator. 
The simulation's visualization can be exhibited online to help users better understand the driving environment and ego's behavior.}
\label{fig: I-SIM}
\vspace{-1.3em}
\end{figure}

We construct two driving scenario sets to benchmark the effectiveness and generalization ability of the proposed algorithm, respectively. 
\begin{itemize}
\item {Small-scale highly interactive scenarios:} Similarly with \cite{rhinehart2021contingencies}, we select 8 highly interactive scenarios including unprotected left turn, car following, and cut-in in dense traffic. We demonstrate the effectiveness of our algorithm like the traditional RL style, in which training and testing are in the same environments.
\item  {Large-scale interactive scenarios:} To further investigate the scalable and generalization abilities of our method like\cite{huang2023differentiable}, we split the remaining 632 meaningful interactive scenarios as 476 training and 156 testing.
\end{itemize}

\subsubsection{\textbf{Simulation}}
Directly closing the loop in the real world provides the best realism but is unsafe. In general, the interactive agent is trained in the simulation environment. Motivated by \cite{bronstein2022hierarchical},\cite{gulino2023waymax}, the log-replay scenarios built upon the offline dataset are adopted for training and evaluation. 
Specifically, we use the I-SIM\cite{zhang2022trajgen}, which is an open-sourced log-replay simulator for the INTERACTION dataset. The workflow is illustrated in Fig. \ref{fig: I-SIM}, where the simulator runs in a docker container and it can extract static road information from the map file in OSM form and dynamic vehicle information from several recorded track files in OSV form. Vehicle kinematics in I-SIM are modeled using a 2D bicycle model for the ego vehicles. 
The behaviors of other road users are replayed from the log record. 
The simulator provides different kinds of functions to calculate features like vehicles’ distance and collision flag, which are essential to get observations and rewards in the designed POMDP. 
Connections between the agent and the simulator are realized through a client interface with socket communication. 
I-SIM also provides visualization tools to display the simulation. The simulation and control frequency are set to 10 HZ.

In the training phase, we reset the environment only when the ego vehicle reaches its logged destination or maximum time duration, here we define the ego's maximum time duration as equal to its actual existing time in the log record. 
Collision is not considered as a terminal condition since the trajectory prediction task can earn profit from more complete episodes in the early stage of the training. 
However, in the evaluation phase, collisions will end the episode to calculate the metrics.
For the closed-loop training, we randomly select one from 8 scenarios when a new episode begins in the small-scale experiment. In the large-scale experiment, considering the difficulties vary between different scenarios since they are highly diverse in track shape and traffic density, the random selection strategy may cause an unfair comparison of the training curves, therefore we select the ego vehicle in a fixed order.

\begin{figure*}[htbp]
    \centering
    \includegraphics[width=18cm]{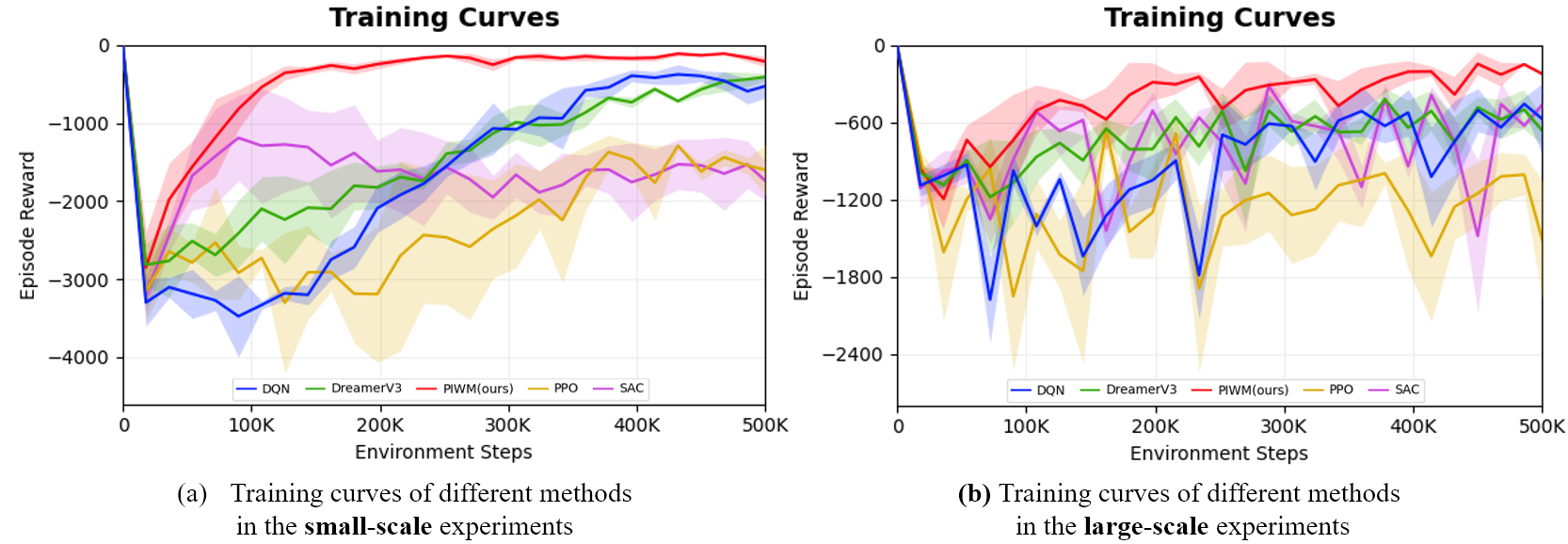}
\caption{Training curves of our proposed method PIWM and other learning-based baselines. PIWM shows superior performance in terms of sample efficiency and final performance on both benchmark scenarios.}
\label{fig: compare curves}
\vspace{-1.3em}
\end{figure*}

\subsubsection{\textbf{Observation Space}}
As discussed in \ref{section: problem formulation}, we suppose the observation $x_t$ at time $t$ contains vehicles' historical positions and headings. More specifically, we treat vehicles' observations as vectorized historical $2s$ trajectories with a fixed temporal interval of $0.1s$. Consider that the raw historical trajectory of a vehicle at time $t$ can be represented by a list of 2D coordinates $[\bold{d}_{t-19}, \bold{d}_{t-18},...,\bold{d}_{t}]$, where $\bold{d}_{i}, (i=t-19,...,t)$ is the 2D coordinate, the observation can be represented by a list of vectors $[\bold{v}_{t-18}, \bold{v}_{t-17},...,\bold{v}_{t}]$, for each element $\bold{v}_{i}, (i=t-18,...,t)$ is given by:
\begin{equation}
\begin{split}
\begin{array}{ll}
\bold{v}_{i} = [\bold{d}_{i-1}, \bold{d}_i, \bold{yaw}_i]
\end{array}
\end{split}
\end{equation}
where $\bold{yaw}_i, (i=t-18,...,t)$ is the vehicle's heading angle. The vectorization\cite{gao2020vectornet} of the trajectories can better represent the dynamics of vehicles without extra information or further modification to the raw trajectories. We introduce the headings as supplement information for vehicles' attitudes.
For the ground truth of prediction trajectories $y_t=\{y_t^{ego}, y_t^{VDI}\}$, we treat them as lists of 2D coordinates, the prediction horizon is set to $2s$ in the following experiments.

The ego's vehicle detective range is set to $30$m in its directly behind and $60$m otherwise. 
Without loss of generality, we consider $N=10$ surrounding vehicles that are closest to the ego vehicle as social vehicles, 
they are sorted by the distances to the ego vehicle, in which the closest $N_{d}=5$ vehicles are VDI and the other $N_{p}=5$ vehicles are VPI.
The observations are zero-padded if there are not enough surrounding vehicles.
Here the VDI-VPI are in equal proportion, and our empirical experimentation results suggest that the proportion only has a minor impact on final performance.
Note that the observation $x_t$ and ground truth future $y_t$ are all transferred into the ego's coordinate frame to make them invariant to the locations.

\subsubsection{\textbf{Action Space}}
We focus on the longitudinal decision task in this paper and design the action space as a fixed-size set of discrete desired speeds, i.e. $a \in [0,3,6,9] \ m/s$. 

The ego vehicle moves along its recorded path represented as a sequence of equidistant waypoints. We adopt two PID controllers to finish the low-level control task, in which the longitudinal PID controller $\texttt{PID}_\texttt{lon}$ brings the ego to the desired speed, and the lateral PID controller $\texttt{PID}_\texttt{lat}$ controls the ego's front wheel angle to track the future path.

\subsubsection{\textbf{Reward Function}}
We use the following reward function to balance safety and efficiency in our experiments:
\begin{equation}
\begin{split}
    r = r_{collision}+r_{speed}+r_{step} \\
\end{split}
\end{equation}

\begin{itemize}
\item $r_{collision}$: Every time the agent collides with other vehicles, a negative reward of -30*(1+$v\_norm$) is given.
\item $r_{speed}$: The agent gets a positive reward of 0.3*$v\_norm$ for every step.
\item $r_{step}$: The agent gets a fixed negative reward of -0.3 for every time step.
\end{itemize}
where $v\_norm$ is the ego vehicle's normalized speed value. 
Adjusting $r_{collision}$'s scale with the ego's speed encourages agents to avoid active collisions. 
The speed reward $r_{speed}$ is used to encourage the ego to drive efficiently. We also set a constant negative reward $r_{step}$ to prevent the ego from standing still. 
Note that in this work, the designed reward function mainly focuses on safety and efficiency, more efforts can be made in reward engineering if other aspects of driving are considered, such as comfort and courtesy.

\subsection{Baseline Methods}
We compare our proposed method with the following baselines. 
Note that baseline methods learn from a scene-level representation. Hence the observations of the ego vehicle and $N$ surrounding social vehicles are sorted by their Euclidean distances to the ego and then concatenate together as a whole to represent the driving scene.

\subsubsection{Random} Ego vehicles randomly select one action from the discrete action space every step. The collision rate of the random policy can well reflect the interactivity of a log-replay scenario.

\subsubsection{Deep Q Learning (DQN)\cite{mnih2015human}} DQN is the most well-known model-free and off-policy RL method, which is suitable for the driving task with discrete action space. 

\subsubsection{Proximal Policy Optimization (PPO)\cite{schulman2017proximal}} PPO is a typical model-free and on-policy RL method. It stabilizes the training progress by limiting the divergence between new and old policies with a clipped surrogate objective function.

\subsubsection{Soft Actor-Critic (SAC)\cite{haarnoja2018soft}} SAC is the SOTA model-free and off-policy RL method within the Maximum Entropy RL framework. 
We deploy its discrete version\cite{christodoulou2019soft} in this paper to fit the action space. 

\subsubsection{DreamerV3\cite{hafner2023dreamerv3}}DreamerV3 is the SOTA model-based RL method, which learns a differentiable world model and utilizes it to train the actor-critic fully in imagination. 

To ensure fairness in comparison, we keep the same-level model size for the model-based baseline ($9.5M$ and $9.8M$ parameters for DreamerV3 and PIWM, respectively). For model-free baselines, we design their encoders, and actor-critics (only critic for DQN) have the same structure as in DreamerV3.

\begin{table}[htbp]
    \caption{
    Evaluation results compared with baselines. The upper and lower parts are the results of the small- and large-scale experiments, respectively. Each experiment is conducted across 3 runs.
    }
    \label{tab: evaluation results}
    \renewcommand{\arraystretch}{1.4}
    \small
    \centering
    \vspace{0.10cm} 
    \resizebox{\linewidth}{!}{
    \begin{tabular}{c|c|cc|c}
    \toprule
    \multicolumn{1}{c|}{\multirow{2}{*}{\textbf{Methods}}} & \multicolumn{1}{c|}{\textbf{Success}} & \multicolumn{2}{c|}{\textbf{Failure Rate(\%)} $\downarrow$} & \multicolumn{1}{c}{\textbf{Average Comple-}} \\
    \cline {3 - 4} & \multicolumn{1}{c|}{\textbf{Rate (\%)} $\uparrow$} & \multicolumn{1}{c}{\textbf{Collision}} & \multicolumn{1}{c|}{\textbf{Time-exceed}} & \textbf{tion Ratio (\%)} $\uparrow$ \\ 
    \midrule
    Random  &  0.00$\pm$0.00  &  100.00$\pm$0.00 & 0.00$\pm$0.00  &  27.77$\pm$0.85    \\ 
    DQN  &  29.17$\pm$5.89  &  45.83$\pm$5.89  &  25.00$\pm$0.00  &  66.74$\pm$4.81    \\ 
    PPO  &  0.00$\pm$0.00  &  50.00$\pm$4.08  &  50.00$\pm$4.08  &  8.56$\pm$1.87    \\ 
    SAC  &  9.17$\pm$2.36  &  90.83$\pm$2.36  &  0.00$\pm$0.00 &  30.32$\pm$0.78    \\ 
    DreamerV3  &  \underline{40.00$\pm$3.54}  &  \underline{38.33$\pm$11.96}  &  \underline{21.67$\pm$8.50}  &  \underline{71.83$\pm$3.38}    \\ 
    PIWM(ours)  &  \textbf{85.00$\pm$4.08}  &  \textbf{14.17$\pm$5.14}  &  \textbf{0.83$\pm$1.18}  &  \textbf{91.54$\pm$2.45}    \\ 
    \midrule
    Random  &  41.67$\pm$1.05  &  46.79$\pm$2.28  &  11.54$\pm$1.57  &71.16$\pm$0.48    \\ 
    DQN  &  42.52$\pm$6.59  &  34.40$\pm$3.48  &  23.08$\pm$9.94  &  74.29$\pm$2.17    \\ 
    PPO  &  33.12$\pm$10.76  &  43.80$\pm$1.84  &  23.08$\pm$12.37  &  62.99$\pm$9.99    \\ 
    SAC  &  45.94$\pm$3.15  &  38.46$\pm$2.28  &  15.60$\pm$3.93  &  76.21$\pm$1.04    \\ 
    DreamerV3  &  \underline{55.98$\pm$1.21}  &  \underline{32.05$\pm$1.05}  &  \underline{11.97$\pm$1.60}  &  \underline{78.46$\pm$0.95}    \\ 
    PIWM(ours)  &  \textbf{74.79$\pm$2.58}  &  \textbf{23.93$\pm$3.07}  &  \textbf{1.28$\pm$0.91}  &  \textbf{88.07$\pm$1.29}    \\ 
    \bottomrule
    \end{tabular}}
\vspace{-1.3em}
\end{table}

\subsection{Metrics}
We evaluate agents from multiple perspectives using the following metrics:

\subsubsection{Episode Reward}
It is a commonly used metric in RL and represents the sum of rewards in an episode.
\subsubsection{Success Rate}
The percentages of episodes where the ego vehicle completes at least $90\%$ of its track without any collisions. 
\subsubsection{Failure Rate}
The percentages of episodes where the ego vehicle doesn't succeed. Specifically, the failure episodes can be divided into two conditions: 
\begin{itemize}
\item Collision, in which the ego vehicle ends the episode with a collision event.
\item Time-exceed, in which no collision event occurs but the ego vehicle fails to finish at least $90\%$ of its track on time.
\end{itemize}
\subsubsection{Average Completion Ratio} 
The average ratio of the track's length completed by the ego vehicle to the total track's length before the episode ends.

\subsection{Performance Comparison}
We compare the proposed method with competitive baselines from the training and evaluation phases on the small- and large-scale benchmark scenarios, respectively. The learning-based methods are all trained three times with INTEL i7-7800X CPU and NVIDIA RTX 2080Ti GPU.

\subsubsection{\textbf{Our method performs better on both benchmark scenarios during the training phase}}

Fig. \ref{fig: compare curves} presents the training curves of the proposed method and other learning-based baseline methods on two benchmark scenarios. 

\textit{Training results on small-scale highly interactive scenarios}:
As shown in Fig. \ref{fig: compare curves}(a), our method shows a significant advantage in learning efficiency, it takes only $200k$ environment steps to achieve an average performance of -150 and maintain at this level, which is unattainable for baselines even after $500k$ environment steps. 
DreamerV3 is the second-best method considering the learning efficiency and final performance comprehensively. 
For the model-free baselines, DQN is slower than DreamerV3 in terms of learning efficiency but achieves a similar final performance. 
SAC learns surprisingly fast in the early training stage, but it converges to a sub-optimal policy. 
As an on-policy method, PPO suffers from low data efficiency and also converges to a sub-optimal policy.

\textit{Training results on large-scale interactive scenarios}:
Fig. \ref{fig: compare curves}(b) shows our method has significant advantages in both learning efficiency, scalability, and converged performance with smaller fluctuations. 
DQN and SAC's learning processes are unstable compared with DreamerV3, which is likely due to the diversity of the scenarios making it hard for the model-free methods to learn one scalable policy for large-scale scenarios. 
PPO fails in this experiment and its training curve remains flat. 
DreamerV3 learns a scalable policy by training it in the world model\cite{lu2023challenges}, thus stabilizing the training process. 
We will further analyze the performance gap between our method and DreamerV3 in the ablation study.

\begin{figure}[htbp]
    \centering
    \includegraphics[width=9cm]{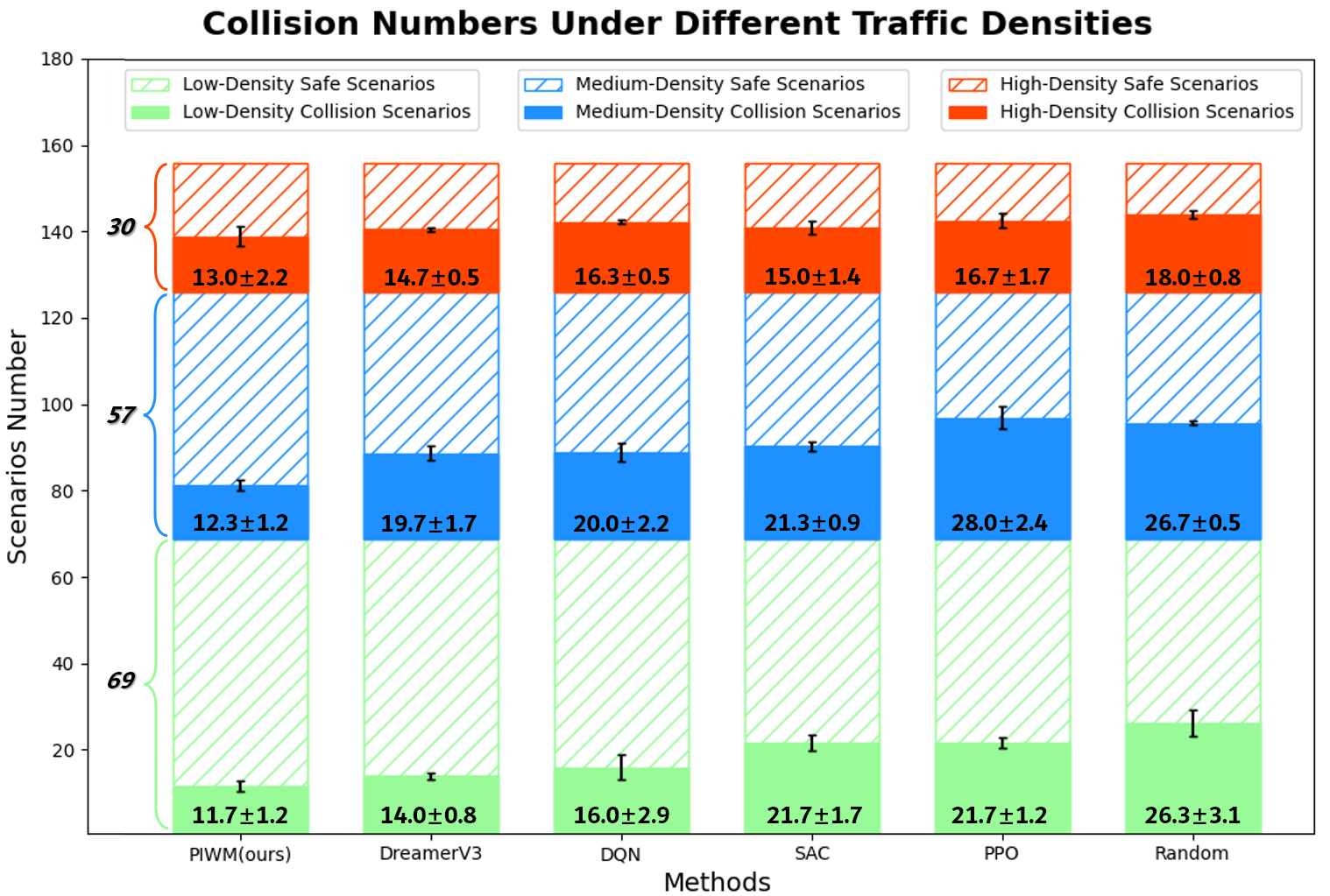}
\caption{
Collision numbers of different methods in the evaluation phase of the large-scale experiments. 
The shaded and solid areas represent safe and collided scenarios, respectively.
The mean and standard deviations of the collision numbers are also given.
}
\label{fig: collisions different densities}
\vspace{-0.5em}
\end{figure}

\subsubsection{\textbf{Our method surpasses all baselines on both benchmark scenarios during the evaluation phase}}
As mentioned in section \ref{section: ES}, we evaluate the models at $500k$ environment steps in the same training scenarios for the small-scale experiments, and 156 new testing scenarios for the large-scale experiments. The evaluation results of metrics except for episode reward are exhibited in Table \ref{tab: evaluation results}. 

\begin{figure*}[htbp]
    \centering
    \includegraphics[width=18cm]{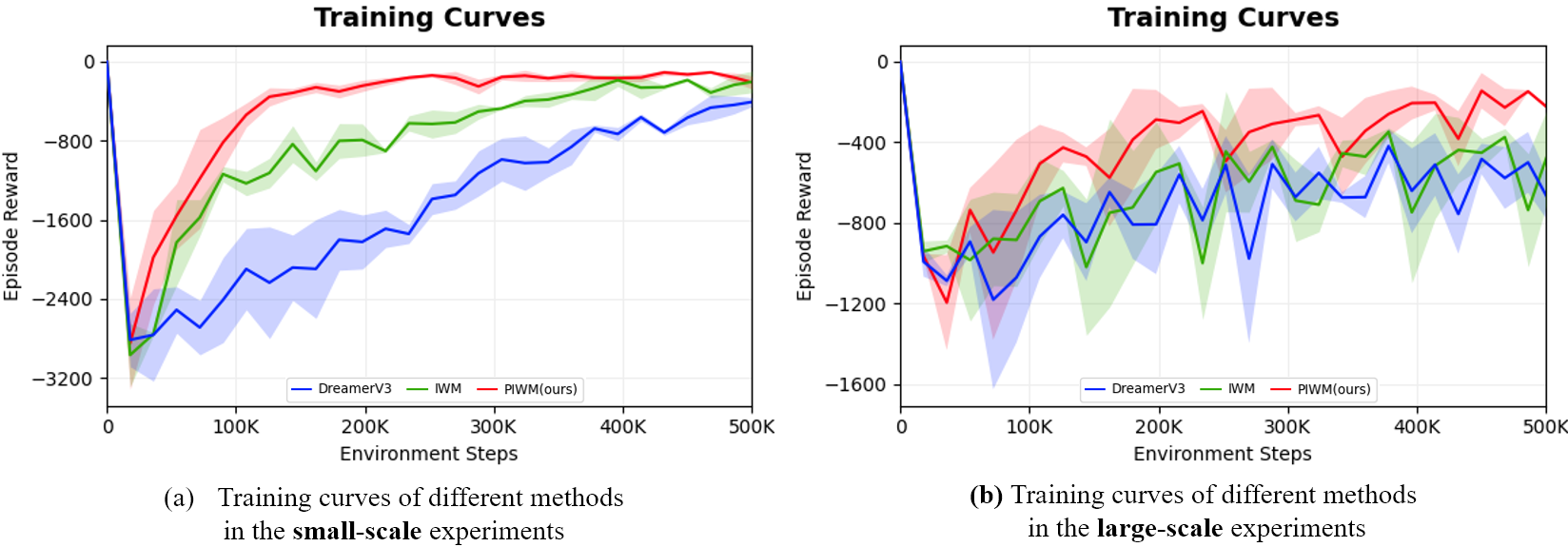}
    \caption{
    Ablation study of the designs of the PIWM.
    Individual modeling helps to improve sample efficiency and final performance significantly in small-scale experiments compared with DreamerV3. 
    Enhanced by interactive prediction, PIWM performs the best of both benchmarks.
    }
\label{fig: ablation on key components}
\vspace{-1.75em}
\end{figure*}

\textit{Evaluation results on small-scale highly interactive scenarios prove the effectiveness of our method}: We evaluate each model for 40 episodes, where each scenario is tested 5 times. 
The table shows that the random policy totally fails in small-scale highly interactive scenarios with $100\%$ collision rate, which indicates the agent should have some level of interactive reasoning ability to navigate safely in such scenarios. 
Our method achieves the highest success rate of $85.00\%$, which dominates all baseline methods, it also has the lowest collision rate of $14.17\%$, the highest completion ratio of $91.54\%$.
DreamerV3 is the second-best method with a $40.00\%$ success rate and a $38.33\%$ collision rate, it has a decent completion ratio of $71.83\%$. 
DQN successfully finishes $29.17\%$ of the test episodes, it learns a conservative policy and drives slowly to avoid collisions, which leads to a relatively low completion ratio.
PPO and SAC are both trapped in local optima,
where PPO learns an overly conservative policy which leads to the worst completion ratio and $0\%$ success rate. SAC on the other hand learns an aggressive policy that has a higher completion ratio compared with PPO but collides with others all the time.

\textit{Evaluation results on large-scale interactive scenarios prove the scalable and generalization abilities of our method}: We test each model for 156 episodes, where each unseen scenario is tested 1 time. The results are similar to the small-scale experiment where our method achieves the best performance in all of the metrics, which indicates that our method owns the scalable ability in large-scale scenarios. 
To further analyze the interactive ability of different methods in large-scale scenarios, we divide the 156 unseen scenarios into 3 traffic density levels. Normally, a scenario with higher traffic density requires more interactive decision-making with sophisticated behaviors and frequent behavior-switching. 
The traffic density level of a scenario is determined by the maximum number of surrounding vehicles $N_{max}$ in the detective range of the \textbf{log-replayed ego vehicle}, the density level is considered as ``low'' when $N_{max} \leq 6$, ``medium'' when $6 \ \textless \ N_{max} \leq 8$, and ``high'' when $8 \ \textless \ N_{max}$. As a comparison, the traffic densities are all ``high'' for small-scale scenarios. We treat the number of collision scenarios as the compare metric.
As shown in Fig. \ref{fig: collisions different densities}. The 156 unseen scenarios consist of 69 low-density scenarios, 57 medium-density scenarios, and 30 high-density scenarios. Our method achieves the lowest number of collision episodes in all three levels of scenarios, showing a sound interactive ability. The random policy can be viewed as a lower limit for avoiding collisions since it doesn't interact with others, it reaches the highest number of collision scenarios in low- and high-density scenarios and the second highest in medium-density scenarios. It is worth noticing that the random policy prevents collision from happening in plenty of scenarios, especially in low- and medium-density scenarios, which indicates that a part of the large-scale scenarios is collision-free and easy to pass.

It should be noted that the success rates of baselines are increased compared with the small-scale experiments, the reason is that not all testing scenarios are highly interactive, as discussed above, the ego vehicle can reach the goal without collisions in some scenarios even with a random policy. Although our method achieves the best performance in unseen scenarios, we notice that its success rate drops compared with the small-scale experiments. It reveals the fact that the generalization ability of model-based RL is still a challenging topic for large-scale diverse driving scenarios.

\subsection{Ablation Study}

\subsubsection{Influence of the designs of individual branches and interactive prediction task to the world model} We conduct ablations on two benchmark scenarios to investigate the effectiveness of the designed individual modeling framework and the interactive prediction representation learning task.

\noindent\textbf{DreamerV3.} World model based on the scene-level modeling and the reconstruction representation learning.

\noindent\textbf{Individual world model (IWM).} To ablate the individual modeling framework, we train a variation of DreamerV3 with the branched networks for individual modeling and the individual reconstruction task for representation learning. For simplicity, we refer to it as IWM. 
Note that due to the absence of interactive relations modeling between vehicles in the latent states, the cross-attention module of IWM considers all surrounding vehicles including VDI and VPI as inputs. 
The reconstruction decoders learn representations by reconstructing both VDI's and VPI's observations since the vehicles' states are independent of each other.

\noindent\textbf{PIWM.} This is our proposed method, where the PIWM combines the individual modeling structure and interactive prediction representation learning.

The training results are shown in Fig. \ref{fig: ablation on key components}.
Compared with DreamerV3, we observe higher learning efficiency and converged performance for IWM in the small-scale experiments, which indicates that the individual modeling does reduce sample complexity and improve decision performance. However, the superiority becomes minor when it comes to large-scale experiments with diverse driving scenarios. 
A possible explanation is that learning latent states solely by reconstructing vehicles' individual observations ignores the interactive relations between vehicles and has trouble capturing vehicles' intentions, which hurts the world model's imagination ability as well as the behavior model's decision performance, this problem becomes severe as the diversity grows. 
As a result, our method with relationship modeling and intention-aware states surpasses IWM in both experiments significantly.

\begin{table}[htbp]
    \caption{{Evaluation results of the designs of the PIWM.
    With individual modeling and interactive prediction representation learning, the success rates of our method are improved.}}
    \label{tab: ablation on prediction}
    \renewcommand{\arraystretch}{1.4}
    \small
    \centering
    \vspace{0.10cm} 
    \begin{tabular}{c|c|c}
    \toprule
    \multicolumn{1}{c|}{\multirow{2}{*}{\textbf{Methods}}} & \multicolumn{2}{c}{\textbf{Success Rate(\%)} $\uparrow$} \\
    \cline {2 - 3} & \multicolumn{1}{c|}{small-scale} & \multicolumn{1}{c}{large-scale} \\
    \midrule
    DreamerV3 & 40.00$\pm$3.54  & 55.98$\pm$1.21 \\
    IWM  &  \underline{50.00$\pm$8.90} & \underline{61.11$\pm$3.20} \\
    PIWM(ours)  &  \textbf{85.00$\pm$4.08} & \textbf{74.79$\pm$2.58} \\
    \bottomrule
    \end{tabular}
\vspace{-0.7em}
\end{table}

The evaluation results on success rate are presented in Table \ref{tab: ablation on prediction}, which supports our statement above.

\subsubsection{Influence of prediction horizons} 
\begin{figure}[htbp]
    \centering
    \includegraphics[width=8.5cm]{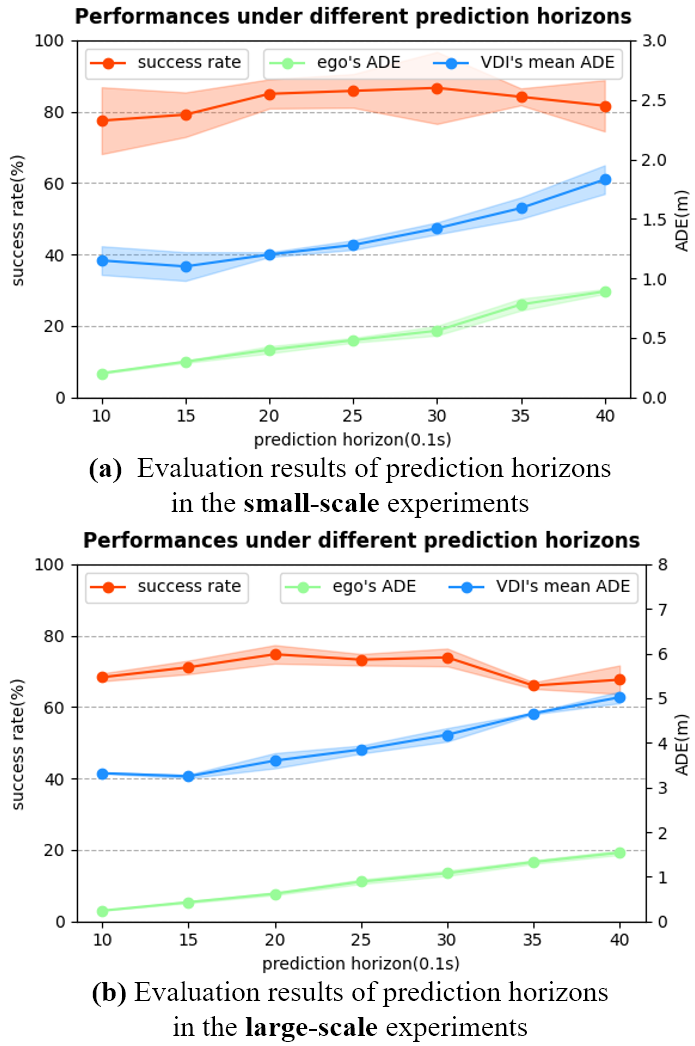}
    \caption{Ablation study of prediction horizons on both benchmark scenarios. 
    As a result, the prediction horizon is set to $20$ to achieve the best trade-off between decision performance and prediction accuracy.}
\label{fig: prediction horizen}
\vspace{-1.3em}
\end{figure}

We also investigate the influence of the prediction horizon $H^+$. This work mainly uses the prediction task to learn an informative state representation, like social vehicles' drive intentions or ego vehicle's motion trends. 
On the one hand, as the horizon grows, the states become more informed of vehicles' future, which facilitates the decision performance and helps us understand the scenario. On the other hand, the prediction error grows together with the horizon, which makes the states inaccurate and inevitably degrades the decision performance.
As shown in Fig. \ref{fig: prediction horizen}, we report our model's performance in success rate and prediction accuracy (average distance error, ADE) for both ego and VDI under different horizons. 
Overall, for both small- and large-scale experiments, the prediction error increases as the prediction horizon grows. 
Specifically, when $H^+ \in [20,25,30]$, our method manages to get high success rates and moderate prediction errors. When $H^+ \in [10,15]$, the predictions are too short to show the long-term intentions of vehicles, thus harming the decision performance. However, the success rate deteriorates when $H^+ \in [35,40]$ as the prediction diverges heavily from the ground truth. It shows that $H^+=20$ achieves the best trade-off between success rate and prediction error, which is chosen as the default in the comparison experiments.

The results in Fig. \ref{fig: prediction horizen} suggest a relatively high prediction error for VDI. 
We suppose the first reason is that we neglect map information in this work, which is monumental to constraint prediction trajectories, and the other reason is that we collect training data from the ego's perspective in an online manner, which is limited and biases the data distribution for VDI's prediction. Future work will focus on improving prediction accuracy and training PIWM from real-world datasets in an offline manner.

\subsection{Visualization}
\begin{figure*}[htbp]
    \centering
    \includegraphics[width=18.5cm]{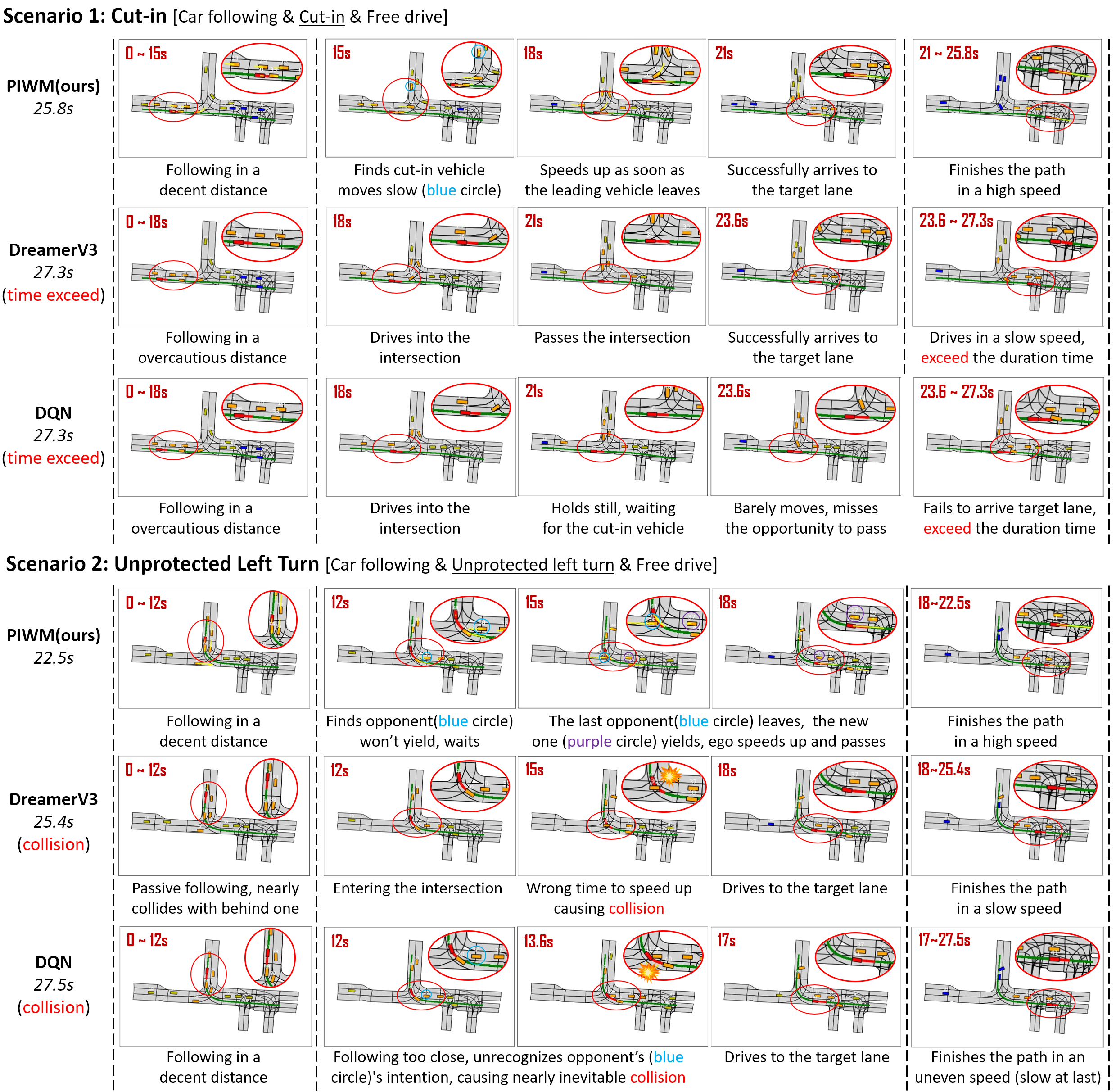}
\caption{Simulated scenarios with sampled key timestamps. Two typical driving scenarios, cut-in and unprotected left turn, are selected, each one of them contains three phases including car following, highly interactive progress (cut-in or left turn), and free drive, which are divided using dashed lines in the figure. For our method, \textcolor{yellow}{\textbf{yellow}} lines are decoded future $2s$ predictions that indicate vehicles' intentions or motion trends, longer lines represent higher speeds. 
The visualization results show that our method can make interactive decisions in complex environments, and the decoded predictions can provide interpretability at some level.
The video version of the simulated scenarios is provided at the project website: \href{}{https://sites.google.com/view/piwm}.
}
\label{fig: visualization}
\end{figure*}

Visualization of our method running in complex driving scenarios can be found in Fig. \ref{fig: visualization}. 
Specifically, we have chosen two typical scenarios for demonstration: cut-in and unprotected left turn. These scenarios require driving agents to make noticeable interactive decisions to navigate both safely and efficiently.
Compared with the SOTA model-based method DreamerV3\cite{hafner2023dreamerv3} and popular model-free method DQN\cite{mnih2015human}, 
they both fail in the typical scenarios. In the cut-in scenario, they adopt overly cautious policies resulting in exceeding the task's time duration. In the unprotected left turn scenario, they tend to collide with other vehicles.
However, our method, PIWM, succeeds in both scenarios, attributable to its ability to identify the intentions of VDI and make interactive decisions. 
Furthermore, for our method, the predicted trajectories of ego and VDI are illustrated as well, which supports that the latent states do capture the interactive relations between vehicles and gain the intentions or motion trends to a certain degree through the interactive prediction task, thereby enhancing the interpretability of our method.

\section{Conclusion}
In this paper, we present a novel model-based RL method for autonomous driving tasks. 
The core idea is the predictive individual world model (PIWM), which learns to encode driving scenarios from an individual perspective, and captures vehicles' inner dependencies and future intentions by incorporating interactive prediction as the representation learning task. 
Therefore the driving policy benefits from the explicit modeling of vehicles and informative latent states. 
Validations on challenging driving scenarios imported from real-world dataset demonstrates our method significantly outperforms the SOTA model-based RL method DreamerV3 and several typical model-free RL methods in terms of learning efficiency and decision performance. Illustrations of the driving scenarios indicate that our method can make interactive maneuvers, and further improve interpretability by decoding prediction trajectories. 
While PIWM achieves superior performance in the experiments, it should be noted that agents are trained and evaluated with non-reactive background traffic. Designing reactive traffic scenarios with human-like agent behaviors is an important research direction for autonomous driving, and investigating RL algorithms in the simulator with both non-reactive and reactive traffic scenarios should be further considered.
Regarding the reward function design, we focus mainly on safety and efficiency, other facets like comfort and courtesy are omitted.
Future directions of our work include expanding PIWM into larger and reactive traffic scenarios, developing a more reasonable reward function considering social interactions, and designing a learnable selection method for vehicles of interest.

\bibliographystyle{IEEEtran}
\bibliography{ref} 


\begin{IEEEbiography}
[{\includegraphics[width=1in,height=1.3in,clip,keepaspectratio]{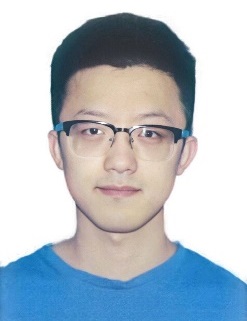}}]
{Yinfeng Gao} received the B.S. degree from the University of Science and Technology Beijing, Beijing, China, in 2019. He is currently pursuing a Ph.D. degree with the School of Automation and Electrical Engineering, University of Science and Technology Beijing, Beijing, China. His research interests include deep reinforcement learning and decision-making with learning-based methods for autonomous driving.
\end{IEEEbiography}

\begin{IEEEbiography}
[{\includegraphics[width=1in,height=1.3in,clip,keepaspectratio]{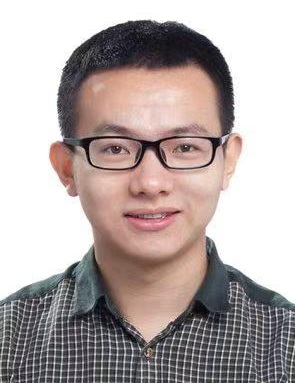}}]
{Qichao Zhang} received the B.S. degree from Northeastern Electric Power University, Jilin, China, the M.S. degree from Northeast University, Shenyang, China, and the Ph.D. degree from the Institute of Automation, Chinese Academy of Sciences, Beijing, China, in 2012, 2014, and 2017, respectively. He is now an associate professor with the Institute of Automation, Chinese Academy of Sciences, and also with the University of Chinese Academy of Sciences, China. His current research interests include reinforcement learning and autonomous driving.
\end{IEEEbiography}

\begin{IEEEbiography}
[{\includegraphics[width=1in,height=1.3in,clip,keepaspectratio]{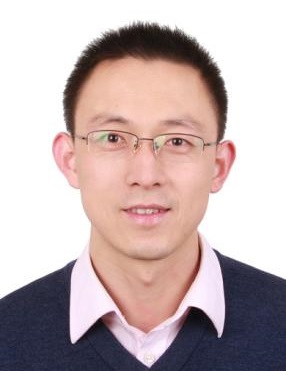}}]
{Dawei Ding} received the B.E. degree from the Ocean University of China, Qingdao, P. R. China, in 2003 and the Ph.D. degree in control theory and engineering from Northeastern University, Shenyang, P. R. China, in 2010. He is currently a Professor with the School of Automation and Electrical Engineering, University of Science and Technology Beijing, Beijing, China. His research interests include robust control and filtering, multi-agent systems, and cyber-physical systems.
\end{IEEEbiography}

\begin{IEEEbiography}
[{\includegraphics[width=1in,height=1.3in,clip,keepaspectratio]{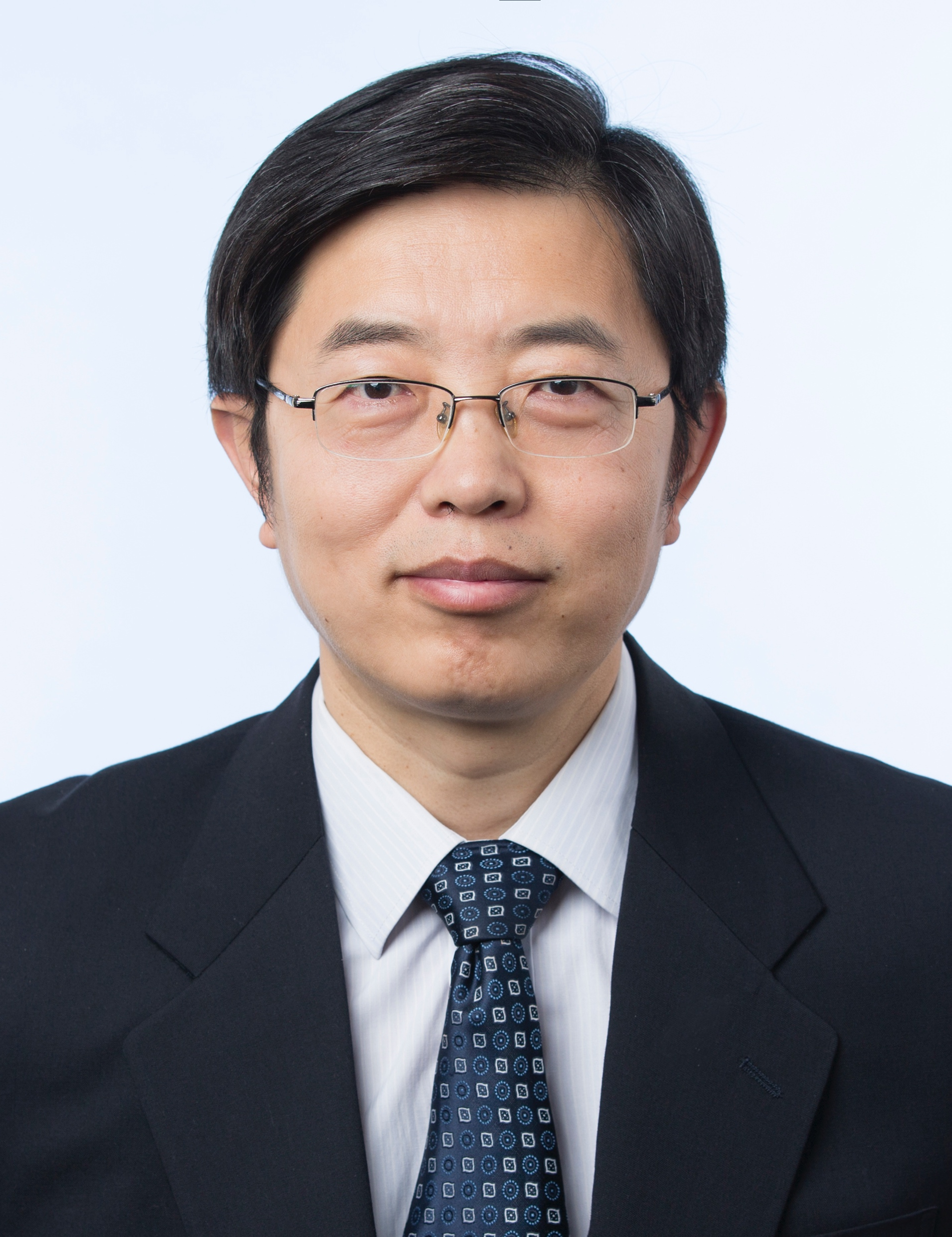}}]
{Dongbin Zhao} (M'06-SM'10-F'20) received the B.S., M.S., Ph.D. degrees from Harbin Institute of Technology, Harbin, China, in 1994, 1996, and 2000 respectively. He is now a professor with Institute of Automation, Chinese Academy of Sciences, and also with the University of Chinese Academy of Sciences, China. His current research interests are in the area of deep reinforcement learning, computational intelligence, autonomous driving, game artificial intelligence, robotics, etc.
\end{IEEEbiography}

\end{document}